\documentclass[conference]{IEEEtran}
\pdfoutput=1
\usepackage{epsfig}
\usepackage{amsmath}
\usepackage{pdflscape}
\usepackage{graphicx}
\usepackage{float}
\usepackage{xcolor}
\usepackage{booktabs}
\usepackage{soul}

\usepackage{tikz}
\usepackage{tikz-cd}
\usepackage{pgfplots}
\pgfplotsset{compat=1.18}

   \everymath=\expandafter{\the\everymath\displaystyle}
   
   \makeatletter\@ifpackageloaded{underscore}{}{\usepackage[strings]{underscore}}\makeatother

\begin{document}

\title{A Vision-Based Closed-Form Solution for Measuring the Rotation Rate of an Object by Tracking One Point}

\author{
\IEEEauthorblockN{Daniel Raviv}
\IEEEauthorblockA{College of Engineering and\\ Computer Science\\
	Florida Atlantic University\\
	Boca Raton, Florida 33431\\
	Email: ravivd@fau.edu}
\and 
\IEEEauthorblockN{Juan D. Yepes}
\IEEEauthorblockA{College of Engineering and\\ Computer Science\\
Florida Atlantic University\\
Boca Raton, Florida 33431\\
Email: jyepes@fau.edu}
\and
\IEEEauthorblockN{Eiki M. Martinson}
\IEEEauthorblockA{College of Engineering and\\ Computer Science\\
Florida Atlantic University\\
Boca Raton, Florida 33431\\
Email: emartins@fau.edu}
}

\maketitle

\begin{abstract}

We demonstrate that, under orthographic projection and with a camera fixated on a point located on a rigid body, the rotation of that body can be analytically obtained by tracking only one other feature in the image. With some exceptions, any tracked point, regardless of its location on the body, yields the same value of the instantaneous rotation rate. 

The proposed method is independent of the shape of the 3D object and does not require \textit{a priori} knowledge about the scene. This algorithm is suited for parallel processing and can achieve segmentation of the scene by distinguishing points that do not belong to the same rigid body, simply because they do not produce the same value of the rotation. This paper presents an analytical derivation, simulation results, and results from real video data.

\end{abstract}


\IEEEpeerreviewmaketitle

\section{Introduction}

This paper presents a novel, simple, analytical closed-form expression for obtaining the rotation rate of a rigid object using a single point tracked over a sequence of images.
When the camera is stationary relative to the object, we demonstrate that, under the assumption of orthographic projection, the object's rotation around the fixation point can be calculated by tracking only one point in the image at three time instants. We show an expression for the rotation rate which is independent of the choice of fixation point on the rigid body and also independent of the choice of the tracked point. In other words, for a specific time instant, the closed form analytical solution we present will always result in the same value, regardless of the point's location on the body.

Obviously all points in a rotating rigid body will have the same angular velocity around the axis of rotation. The rotation rate, which we denote as $\omega$, is therefore an example of a \textit{visual motion invariant} \cite{raviv1993invariants}, an extension into time (by comparing, in this method, sequential frames) of Gibson's idea of the visual invariant \cite{gibson1967new}.

We outline the coordinate system, motion assumptions, advantages and limitations, and derivations. The method is independent of the 3D geometry of the object and does not require \textit{a priori} knowledge about the scene; there is no need for 3D reconstruction of the environment. It is suited for parallel processing and can recognize points that do not result in the same value for $\omega$ as belonging to another moving object, potentially providing a new means of motion segmentation.

We also present simulation results and results obtained from
real-world sequences of images. The non-ideal, non-orthographic nature
of real cameras results in errors computing the rotation rate separate
from and in addition to measurement errors.

The main contribution of this paper: by tracking any point relative to
an arbitrary fixation point on the same rigid body, we obtain a
closed-form solution for the rate of rotation of that body.

\subsection{Related Literature}
The body of literature addressing the visual measurement of rotation rates of objects is quite sparse.

One solution was proposed in \cite{raviv1994visual}. Also assuming orthographic projection, they developed a closed-form solution for the perceived rate of rotation during fixated motion by using the angles (and their first derivatives) of just two image points at two time instants.

The visual processing of rotary motion from a top-down perspective is demonstrated in \cite{werkhoven1991visual}. However, these methods are not applicable to the side view of a rotating object. In addition, they claim that the perceived angular velocity depends strongly on the distance of measured points in the rigid body from the center of rotation. The concept of perceived rotation was explored in \cite{koenderink1987facts} but specific measurement methods were not provided.

In \cite{gardner2019estimating}, a method is proposed for estimating the linear and angular velocities of an object in free flight. This approach involves using two high-speed synchronized stereo cameras to capture images of the object in motion. By extracting features such as lines they estimate the object's velocities. However, this method requires a very specialized setup and is computationally intensive.

The rotation rate can also be determined through egomotion estimation, since the rotation of a rigid body during fixation can be described as the translation and rotation of a camera relative to a stationary rigid body. However, this process is computationally intensive. An overview of some methods and algorithms used for ego motion estimation can be found in \cite{khan2017ego}.

Interestingly, most computer vision literature on visual invariants primarily addresses those derived from still images, as extensively discussed in \cite{flusser2006moment}, \cite{pizlo1994theory}, \cite{weiss1993geometric}, and \cite{zoccolan2015invariant}.

The approach discussed in this paper is \textit{not} applicable to still images; it is based on changes observed in consecutive images  \cite{raviv1993invariants}. More recent work on visual motion invariants can be found in \cite{yepes2023time} and \cite{yepes2023invariant}.

Only a handful of researchers, primarily from the field of behavioral psychology, have explored visual motion invariants and their significance in perception. Notable examples include \cite{gibson1967new}, \cite{gibson2014ecological} and \cite{cutting1986perception}. 

Given the focus on visual fixation in this work, we reference several papers that highlight the advantages of using fixation during relative motion.

In \cite{glennerster2001fixation}, it was demonstrated that a fixation-based representation can simplify computational processes. A key contribution of their work is the elimination of the need for an absolute coordinate frame. Additionally, they provide an excellent literature review on the various components of optical flow during fixated motion.

In \cite{daniilidis1997fixation} it was illustrated that fixation simplifies 3D motion estimation by decoupling the motion parameter space. \cite{gallego2017accurate} discusses accurate estimation of angular velocity using an event camera, and spherical retinal flow for a fixating observer is discussed in \cite{thomas1994spherical}.

Fixated motion can also aid in tasks requiring action, such as vision-based control for stabilizing rigid bodies. In \cite{fermuller1993role} it was demonstrated that using flow obtained during fixation over time can simplify navigation tasks.

\cite{raviv1994unified} introduced a quantitative approach to camera fixation that enhances vision-based road following. Their study includes a thorough analysis of spatial aspects during camera fixation, supported by experimental results \cite{raviv1991quantitative}.

The literature on fixation and visual motion invariants in neuroscience is extensive. For instance, \cite{sunkara2015role} explores the role of visual and non-visual cues in creating a rotation-invariant representation of motion direction. \cite{sunkara2016joint} discusses the joint representation of translational and rotational components of optic flow in the parietal cortex. Additionally, \cite{martinez2004role} provides an overview of the role of fixational eye movements in visual perception in nature. However, it is important to note that the direct relationship between visual motion invariants and fixation in neuroscience is beyond the scope of this paper.

\subsection{Advantages}

Our method for obtaining the rotation rate of a rigid body requires tracking only a single point relative to the fixation point. This approach results in a simple closed-form solution which is the same for all 3D points on a rigid body. The fixation and tracking points can be chosen arbitrarily (except for a few singular points, usually those that lead to division by values close to zero) on the object and under orthographic projection will yield the same result.

No 3D reconstruction is needed; the rotation rate is obtained directly from raw visual data and the method is independent of camera resolution. The $\omega$ invariant does not depend on the shape of the 3D object and is also independent of the horizontal distance relative to the fixation point.

The rotation rate is measured using only three time instants. However, it can also be repeated and measured again, as $\omega$ is a slowly changing, continuous, low-bandwidth signal. Doing so can provide a more robust measurement.

The proposed method is suitable for parallel processing, meaning many points can be processed at once; although the results are theoretically the same, derivatives and second derivatives of displacement will be more accurately measured for some points than others. Averaging obtained values of $\omega$ for many points thereby provides a more robust result.

Since all points on the rigid body result in the same value of $\omega$, this approach can also be used to identify moving points that are not part of the rigid object (since values for $\omega$ obtained from these points will differ), enabling the segmentation of objects in a more comprehensive scene as well.

\subsection{Limitations}
There is a binary ambiguity in determining the direction of rotation. While the magnitude of the rotation rate will be correct, it is not possible to discern whether the rotation is clockwise (CW) or counterclockwise (CCW) without assuming in which quadrant about the fixation point the tracked point is located.

When points are in front of or behind the fixation point, large errors may occur due to near singularity, typically resulting from the division by very small values. Discontinuities may also occur due to sign changes or divisions by zero.

The orthographic projection yields exact theoretical values for the rotation rate. However, real cameras use perspective projection, which introduces some errors. These errors are small when the horizontal angle between the fixation point and the projection of the tracked point is small, i.e., when dealing with points within a very narrow horizontal geometric field of view. Additionally, in perspective projection the choice of fixation point matters when computing $\omega$.

With real data the second derivative of a feature's location in the image is usually noisy. However, collective use of results obtained from many points and over time can significantly reduce this error at the cost of additional computation.

\section{Methodology}

\subsection{Assumptions}
We assume the camera is stationary and fixated on a point located on a rigid object, such that the 3D fixation point remains steady and unchanging in the image over time. The rigid object rotates relative to the fixation point, and points on the object rotate accordingly. In addition, we assume that the 3D rotation vector is oriented perpendicular to the optical axis of the camera. This assumption aligns well with Listing's Law, which states that the eye rotates only about axes in the plane orthogonal to its line-of-sight when it is in its ``primary position'' (facing straight ahead) \cite{Wong2004}.

Our analysis assumes an orthographic projection, ignoring distortions due to perspective. This has the consequence that all computations rely solely on the horizontal components of location and flow (and the derivative of flow). Due to this simplification, it is unnecessary to assume that the distance from the 3D fixation point to the camera remains constant over time.

We further assume that the locations of features and the first and second derivatives of these with respect to time are available and can be measured.

\subsection{Cylindrical Coordinate System}
\begin{figure}[h]
	\centering
        \includegraphics[width=8.5cm]{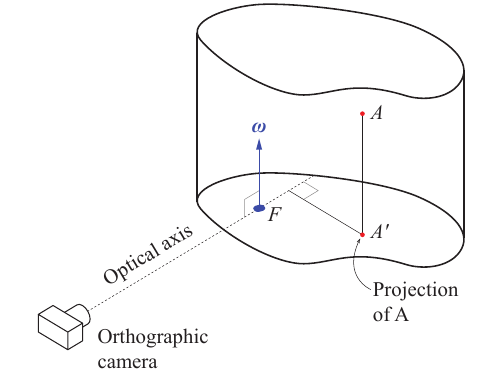}
	\caption{Cylindrical coordinate system}
	\label{cylindrical}
\end{figure}

\begin{figure}[h]
	\centering
        \includegraphics[width=8.5cm]{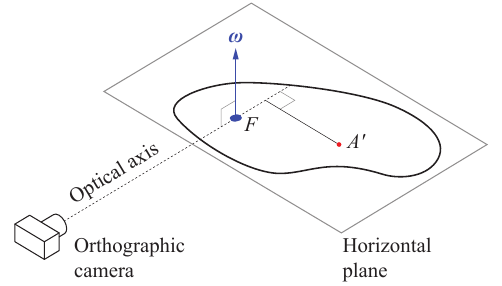}
	\caption{2D plane for simplification of analysis}
	\label{2d-plane}
\end{figure}

In figure \ref{cylindrical}, point $F$ is the fixation point. $A$ is a point in three dimensions on that object, which can be projected onto a 2D plane that contains $F$ and is perpendicular to the rotation vector $\omega$. The projection of $A$ is denoted $A'$. Without loss of generality, we can consider only the horizontal motion of $A'$ on the 2D plane, simplifing analysis, which will be carried out in the plane shown in Figure \ref{2d-plane}.

\begin{figure}[h]
	\centering
        \includegraphics[width=8.5cm]{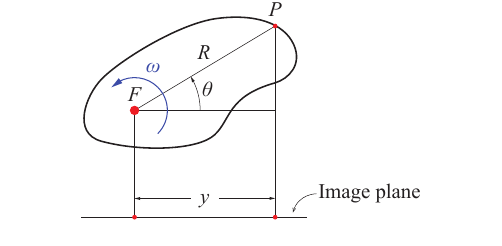}
	\caption{Orthographic projection}
	\label{Orthographic}
\end{figure}

\subsection{Derivations}
Refer to Figure \ref{Orthographic}. Note that, for the following
derivation, the fixation point is chosen at the center of rotation but
this is not necessary; for a proof of the case in which the fixation
point and the center of rotation are not coincident, refer to appendix
\ref{sec:ExtendedProof}. We begin with the expression for the
horizontal displacement $y$ in terms of the angle $\theta$:
\begin{align}
    y &= R \cos{\theta} \label{eq:x_theta}
\end{align}

The physical distance between the fixation point and a point on the rigid body is constant over time, and so the first derivative of $y$ with respect to time $t$ becomes:
\begin{align}
    \frac{\textrm{d}y}{\textrm{d}t} &= R \frac{\textrm{d}}{\textrm{d}t} (\cos{\theta}) \notag\\
    &= R (-\sin{\theta}) \frac{\textrm{d}\theta}{\textrm{d}t} \notag\\
    &= -R \sin{\theta} {\frac{\textrm{d}\theta}{\textrm{d}t}} \label{eq:dy_dt}
\end{align}

where \(\frac{\textrm{d}\theta}{\textrm{d}t}\) is the angular velocity, denoted earlier as $\omega$. Next, we take the derivative with respect to time $t$ of both sides:
\begin{align}
  \frac{\textrm{d}^2y}{\textrm{d}t^2} &= -R \frac{\textrm{d}}{\textrm{d}t} \left[\sin{\theta}\frac{\textrm{d}\theta}{\textrm{d}t}\right] \notag\\
  &= -R \left[\cos{\theta}\frac{\textrm{d}\theta}{\textrm{d}t}\frac{\textrm{d}\theta}{\textrm{d}t} + \sin{\theta}\frac{\textrm{d}^2\theta}{\textrm{d}t^2}\right] \notag
\end{align}
As $\frac{\textrm{d}\theta}{\textrm{d}t} = \omega$ is constant, $\frac{\textrm{d}^2\theta}{\textrm{d}t^2} = 0$ and we can make the following simplifications:
\begin{align}
  \frac{\textrm{d}^2y}{\textrm{d}t^2} = -R  \cos{\theta} \frac{\textrm{d}\theta}{\textrm{d}t}\frac{\textrm{d}\theta}{\textrm{d}t} = -R  \cos{\theta} \left(\omega\right)^2 \label{eq:d2y_dt2}
\end{align}


Now we can substitute equation \ref{eq:x_theta} for $y$ to obtain:
\begin{align}
    \frac{\textrm{d}^2y}{\textrm{d}t^2} &= -y\omega^2 \label{eq:ratio}
\end{align}

Which can be compactly expressed as:
\begin{align}
    \ddot y + y\omega^2 = 0\label{eq:omega}
\end{align}

Therefore, by measuring the second derivative of the displacement $y$ of the point registered by the orthographic camera, we can determine the angular rate of rotation $\omega$. However, as mentioned earlier it is clear that there are two symmetrical solutions for $\omega$, with the same amplitude but different signs. 

Note that in equation \ref{eq:omega}, $\omega^2$ is always non-negative because the ratio between the second derivative of a cosine function over the cosine function itself is non-positive, in other words, $\omega$ is always a real number. Please note that $\omega$ is the magnitude of the vector \(\boldsymbol{w}\) and the sign (based on the vertical direction of \(\boldsymbol{w}\)) is positive for anticlockwise and negative for clockwise.

\section{Results and Analysis}

\subsection{Simulations}
\begin{figure}[h]
	\centering
	{\epsfig{file = 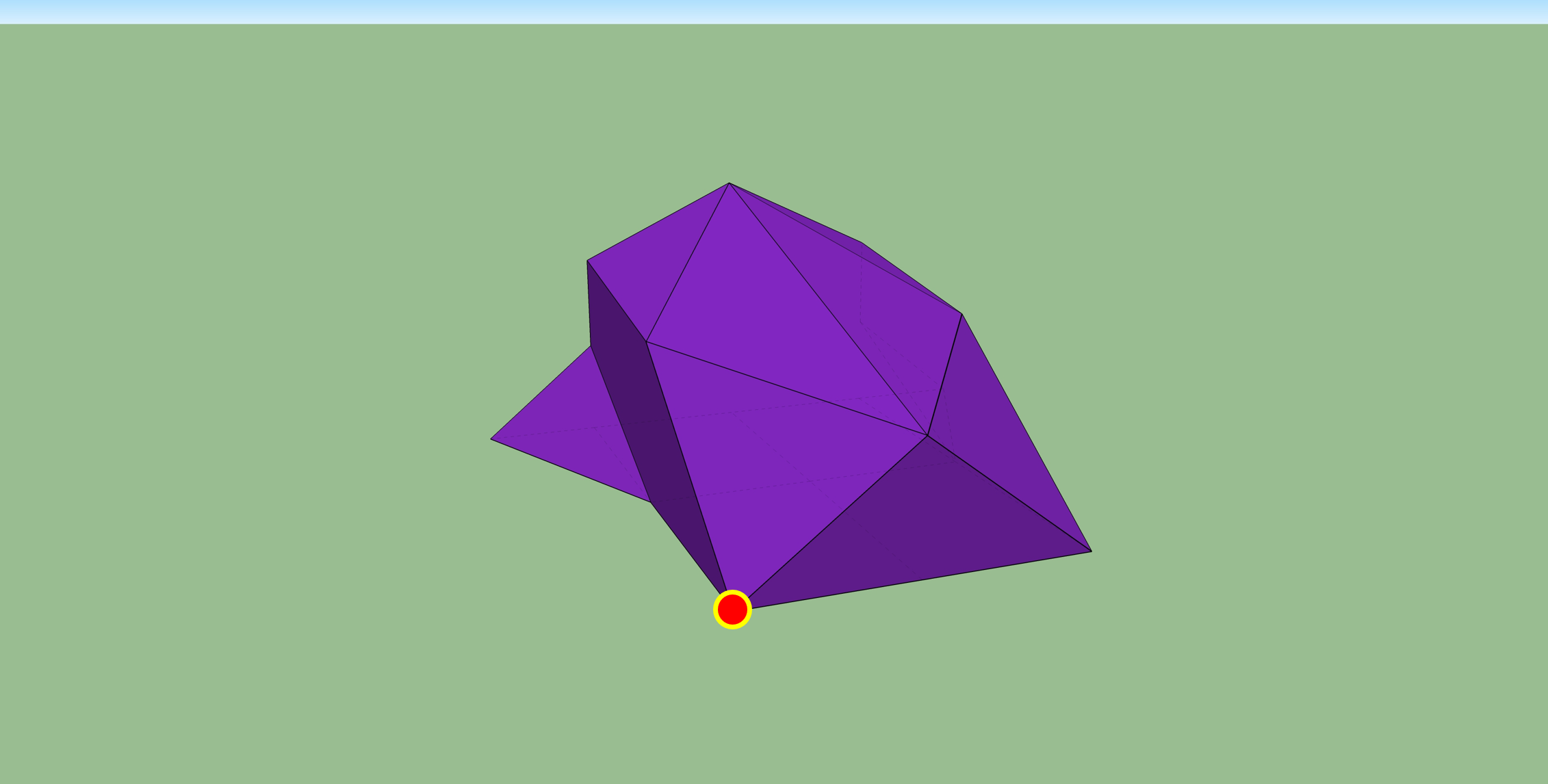, width = 8.5cm}}
	\caption{3D object with a tracked feature point shown in red.}
	\label{tracked}
\end{figure}

We wrote a simulation in Python to visualize the derivation above. Figure \ref{tracked} displays the rotating object; the tracked feature is highlighted with a red circle. The fixation point can be an arbitrary point on the object, but in this case we make it coincident with the axis of rotation for convenience.

\begin{figure}[h]
	\centering
	{\epsfig{file = 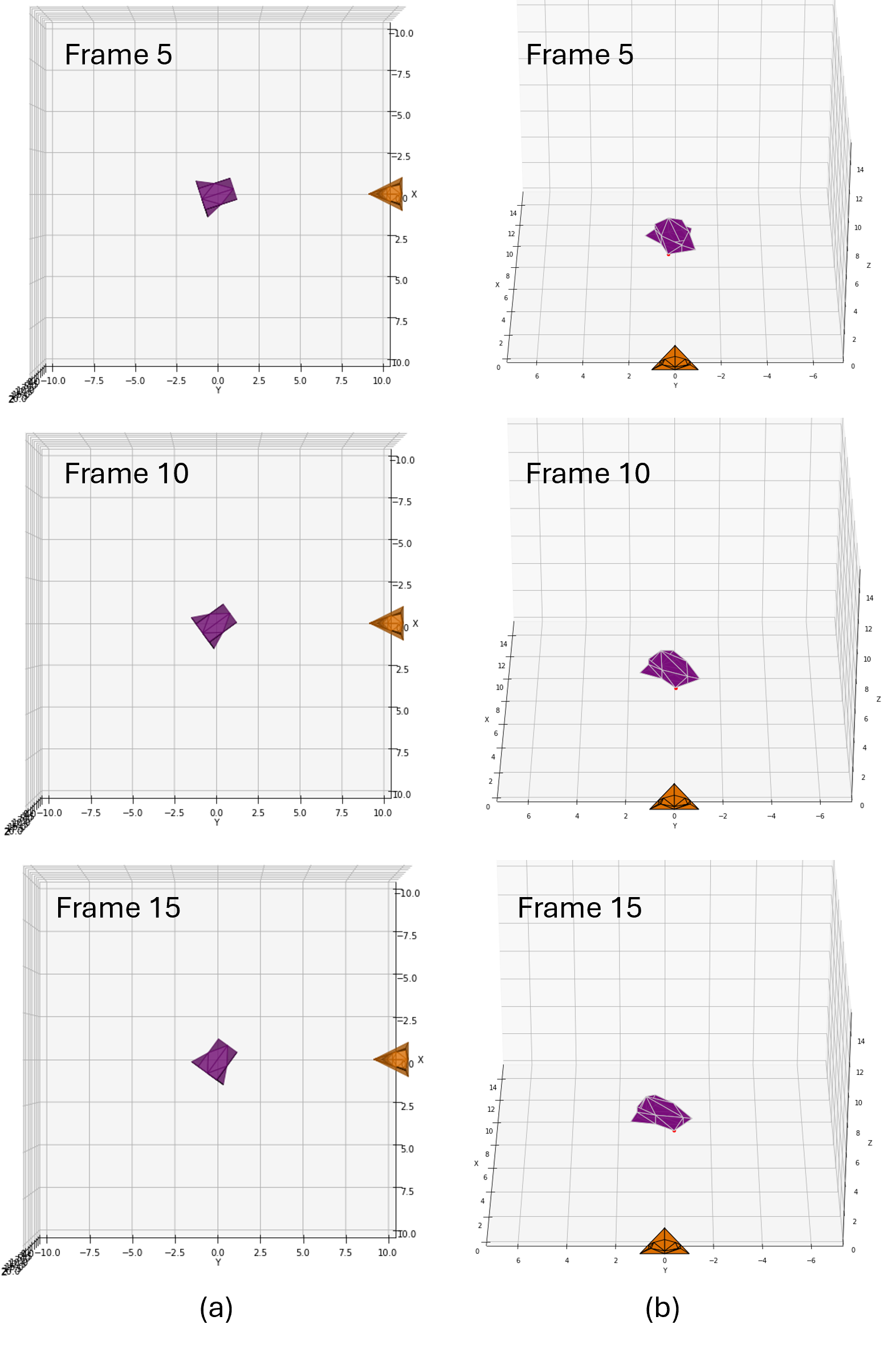, width = 8.5cm}}
	\caption{(a) Top view and (b) slanted view of the object relative to the camera. Camera is depicted as a brown triangle, and the object is drawn in purple.}
	\label{topview}
\end{figure}

Figure \ref{topview} illustrates three sets of two views each of the object relative to the camera. These sets are snapshots of three equally spaced time instants (at frames 5, 10 and 15).

\begin{figure}[h]
	\centering
        \input{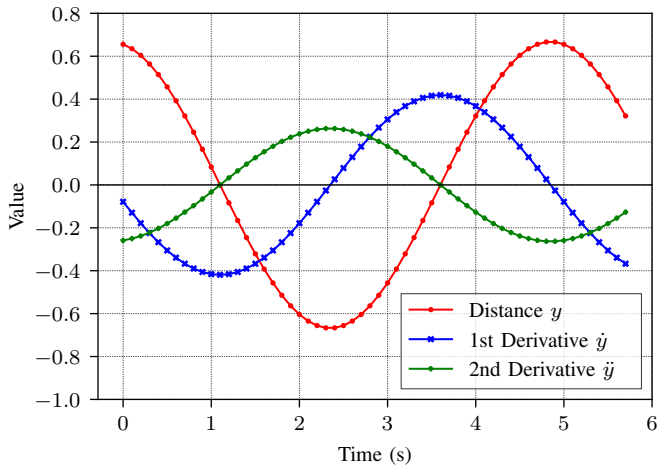}
	\caption{Distance and its first and second derivatives, for a point relative to the fixation point.}
	\label{curves}
\end{figure}

In Figure \ref{curves}, we present results for the projected displacement y, velocity $\textrm{d}y/\textrm{d}t$, and acceleration $\textrm{d}^2y/\textrm{d}t^2$ over time. Notably, the second derivative $\textrm{d}^2y/\textrm{d}t^2$ exhibits an opposite sign to the projected displacement y. Figure \ref{curves} displays results for computed $w^2$ at all time instances. Note the constant resulting value.

\begin{figure}[h]
	\centering
	\input{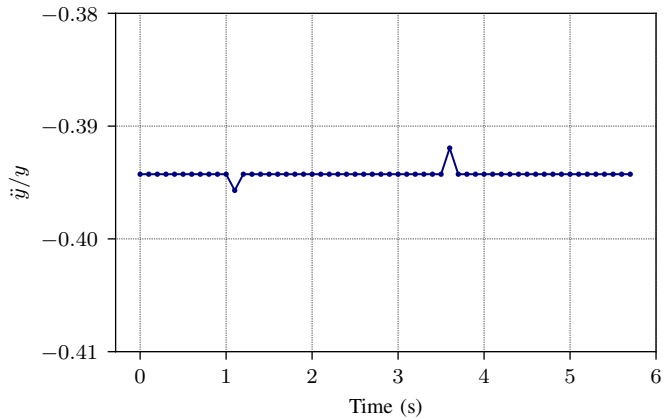}
	\caption{The ratio between the second derivative of the distance over the distance.}
	\label{w-graph}
\end{figure}

Figure \ref{w-graph} show simulation results using the expression in equation \ref{eq:omega}. It shows that the square of the magnitude \(\boldsymbol{w}\) of the angular velocity vector equals the negative ratio between the second derivative of the displacement over the displacement itself. Note the values near time=1.1 sec and time=3.6 sec; they are significantly different because this is close to a singular point.

\subsection{Real data}
\begin{figure}[h]
	\centering
	{\epsfig{file = 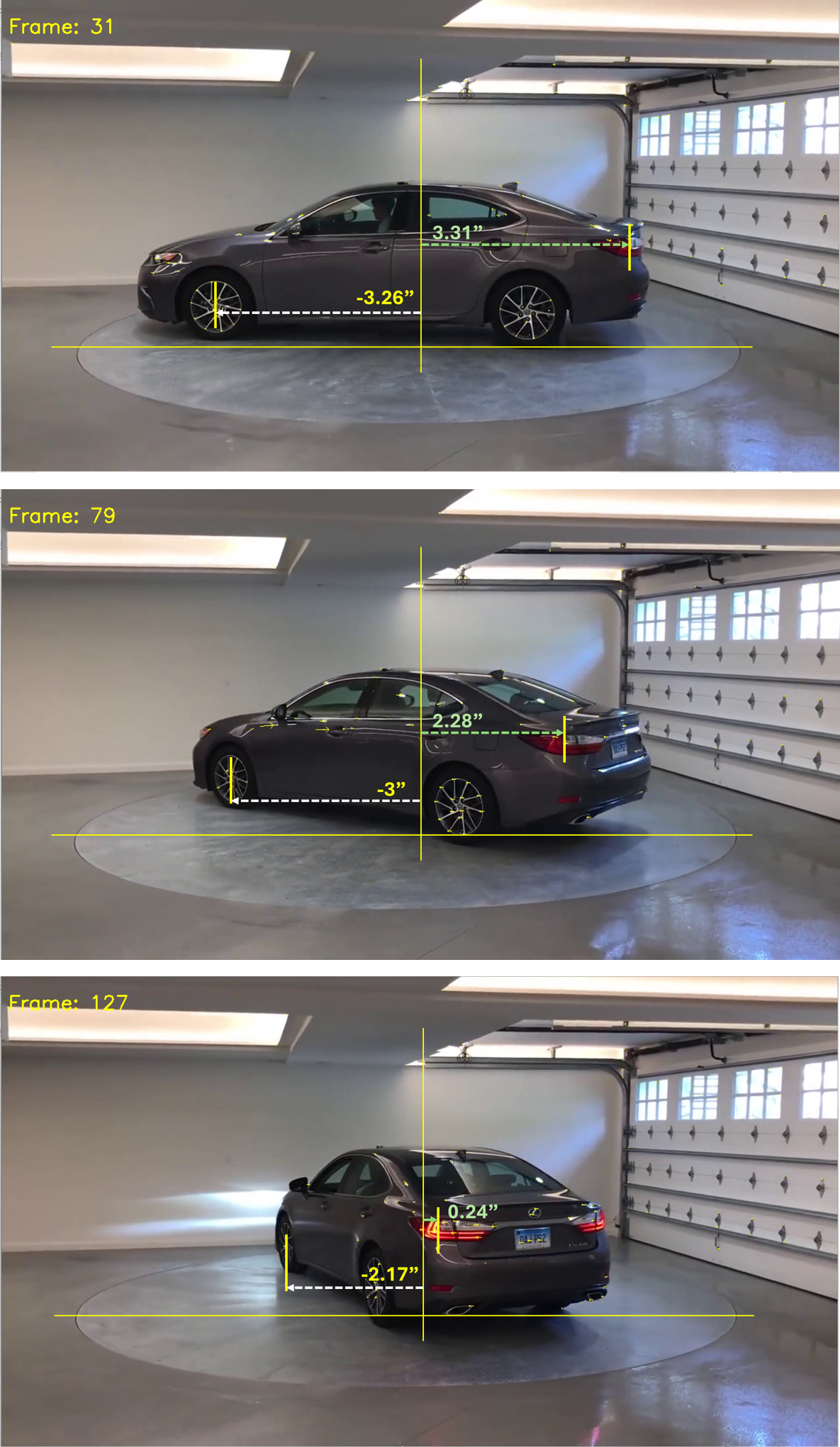, width = 8.0cm}}
	\caption{Three snapshots of a rotating car}
	\label{rotatingCar}
\end{figure}

We analyzed a video featuring a rotating car \cite{VanderHorn2019}. Refer to Figure \ref{rotatingCar} for three snapshots of the video. We manually identified the vertical axis of rotation, shown as the long vertical line in the figure, and tracked two distinct features over time relative to this axis (therefore using it as the fixation point). Applying equation \ref{eq:omega} to these measurements yielded the rotation rate for each tracked point. The images in Figure \ref{rotatingCar} also show short vertical lines that indicate the location of the two tracked features at three video frames (31, 79, and 127).

It should be noted that the point tracking process was performed manually. This was because, at the time of writing, existing optical flow algorithms such as RAFT and Lucas-Kanade, as well as feature identification and tracking algorithms such as SIFT (Scale-Invariant Feature Transform), SURF (Speeded-Up Robust Features), and ORB (Oriented FAST and Rotated BRIEF), did not yield satisfactory results for tracking over time. The ground truth for the rotation rate (0.327 rad/sec) was calculated by measuring the time to complete one rotation.

\begin{table}[!]\centering
  \caption{Measurements and Calculations for Two Features}
  \begin{tabular}{ccccccc}
    \multicolumn{7}{l}{Feature 1:}\\
    \toprule
    Time           & Frame           & $y$              & $\mathrm{d}y/\mathrm{d}t$   & $\mathrm{d}^2y/\mathrm{d}t^2$ & $\omega^2$                      & $\omega$                    \\
    $(\mathrm{s})$ & $(\mathrm{\#})$ & $(\mathrm{in.})$ & $(\mathrm{in.}/\mathrm{s})$ & $(\mathrm{in.}/\mathrm{s}^2)$ & $(\mathrm{rad}^2/\mathrm{s}^2)$ & $(\mathrm{rad}/\mathrm{s})$ \\
    \midrule
    0.00           & 15              & -3.26                                                                                                                                          \\
    0.53           & 31              & -3.26            &  0.00                                                                                                                       \\
    1.07           & 47              & -3.27            & -0.02                       & -0.035                        & -0.01                                                         \\  
    1.60           & 63              & -3.14            &  0.24                       &  0.492                        &  0.16                           & 0.40                        \\   
    2.13           & 79              & -3.00            &  0.26                       &  0.035                        &  0.01                           & 0.11                        \\
    2.67           & 95              & -2.80            &  0.38                       &  0.211                        &  0.08                           & 0.27                        \\
    3.20           & 111             & -2.51            &  0.54                       &  0.316                        &  0.13                           & 0.36                        \\
    3.73           & 127             & -2.17            &  0.64                       &  0.176                        &  0.08                           & 0.28                        \\
    4.27           & 143             & -1.98            &  0.36                       & -0.527                        & -0.27                           &                             \\
    4.80           & 159             & -1.69            &  0.54                       &  0.352                        &  0.21                           & 0.46                        \\
    5.33           & 175                                                                                                                                                              \\
    \bottomrule
    \multicolumn{6}{r}{Average $\omega$, feature 1:}                                                                                                    & 0.31                        \\
  \end{tabular}
  \begin{tabular}{ccccccc}\\
    \multicolumn{7}{l}{Feature 2:}\\
    \toprule
    Time           & Frame           & $y$              & $\mathrm{d}y/\mathrm{d}t$   & $\mathrm{d}^2y/\mathrm{d}t^2$ & $\omega^2$                      & $\omega$                    \\
    $(\mathrm{s})$ & $(\mathrm{\#})$ & $(\mathrm{in.})$ & $(\mathrm{in.}/\mathrm{s})$ & $(\mathrm{in.}/\mathrm{s}^2)$ & $(\mathrm{rad}^2/\mathrm{s}^2)$ & $(\mathrm{rad}/\mathrm{s})$ \\    
    \midrule
    0.00           & 15              &  3.31                                                                                                                                          \\
    0.53           & 31              &  3.31            &  0.00                                                                                                                       \\
    1.07           & 47              &  3.10            & -0.39                       & -0.740                        &  0.24                           & 0.49                        \\    
    1.60           & 63              &  2.79            & -0.58                       & -0.350                        &  0.13                           & 0.35                        \\    
    2.13           & 79              &  2.28            & -0.96                       & -0.700                        &  0.31                           & 0.56                        \\
    2.67           & 95              &  1.66            & -1.16                       & -0.390                        &  0.23                           & 0.48                        \\
    3.20           & 111             &  0.98            & -1.28                       & -0.210                        &  0.22                           & 0.46                        \\
    3.73           & 127             &  0.24            & -1.39                       & -0.210                        &  0.88                           & 0.94                        \\
    4.27           & 143             & -0.67            & -1.71                       & -0.600                        & -0.89                           &                             \\
    4.80           & 159             & -1.42            & -1.41                       &  0.560                        &  0.40                           & 0.63                        \\
    5.33           & 175             & -1.93            & -0.96                       &  0.840                        &  0.44                           & 0.66                        \\
    \bottomrule
    \multicolumn{6}{r}{Average $\omega$, feature 2:}                                                                                                    & 0.57                        \\
    \end{tabular}
    \label{TwoFeaturesResults}
\end{table}

Table \ref{TwoFeaturesResults} summarizes the results obtained for the two points. The sampling rate of the original video is 30 fps. We tracked the two features every 16 frames, starting at frame 15 and ending at frame 175. For each time instant, when possible, we calculated $\omega^2$ and showed only the positive value of $\omega$. Note that sometimes there was no solution due to the negative value of $\omega^2$ because of measurement unavailability and measurement error.

\subsection{Error analysis}  
The error in determining the rotation rate arises from various factors. These include the perspective projection, which deviates from the orthographic method originally used. Perspective introduces divergence and convergence effects that alter the positions of features in the image and their motions across subsequent images. Furthermore, noise inherent in real-world data contributes to errors; in our case this error is very significant since we are using both the first and second derivatives of the position measurements. Additionally, error may stem from camera distortions, the sampling rate during data acquisition, and other less significant effects. 

\section{Conclusion}
In this paper, we demonstrate one method of determining the magnitude of the rotation rate $\omega$ of a rigid object. Assuming orthographic projection, we show that measuring the location of a single point and its second derivative relative to a fixation point is enough to obtain this result. We present mathematical derivations, simulation results, and results from a video of a real rotating object. Note that due to the orthographic projection, any choice of fixation point will result in the same magnitude for $\omega$.

The method presented here has some practical limitations: it requires that the vector of rotation is perpendicular to the line connecting the camera and the fixation point;
tracking a point and its derivative using real images is susceptible to noise; and the method does not provide the sign of the rotation, i.e., whether the measured rotation is clockwise (CW) or counterclockwise (CCW). Finally, it assumes a perfect orthographic projection, impossible with a real camera and approachable only with a very narrow field of view.

Despite all the above limitations, we demonstrate that the method can obtain, even from real data, a reasonable approximation of the rotation rate.

We have tried multiple methods for tracking a point over time to get the location and derivative. However, current methods do not provide good enough tracking results; for the purposes of this paper the points were tracked manually, but point-tracking is a fast-moving area of research where rapid progress can be reasonably anticipated.

The method also demonstrates one way in which fixation can simplify the interpretation of optical flow, as discussed in \cite{glennerster2001fixation}, \cite{daniilidis1997fixation}, and \cite{fermuller1993role}.

The derivation shown in this paper obtains $\omega$ by tracking only one point. We are working also on other methods using multiple points, tracked over time, to achieve better results and robustness.

\section*{Acknowledgment}
The authors would like to thank Michael R. Levine for his continued support of this project. We also thank Dr. Sridhar Kundur for proofreading the paper and providing very useful comments.

\appendices
\section{Determining $\omega$ for unknown center of rotation}
\label{sec:ExtendedProof}

\begin{figure}[h]
	\centering
        \includegraphics[width=8.5cm]{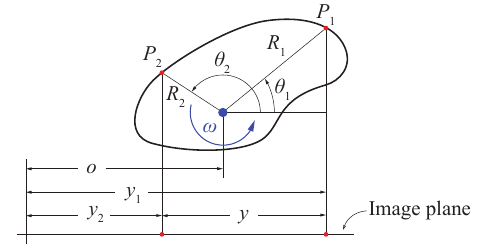}
	\caption{Orthographic projection for two tracked points}
	\label{fig:ExtendedProof}
\end{figure}

Refer to figure \ref{fig:ExtendedProof}. We begin with the expression for the horizontal displacement $y_1$ of point $P_1$:
\begin{align}
    y_1 - o &= R_1 \cos{\theta_1} \label{eq:x1_theta}
\end{align}
$o$ is the unknown (and constant) distance to the center of rotation. Taking the first derivative of both sides therefore yields):
\begin{align}
    \frac{\textrm{d}y_1}{\textrm{d}t} =  -R_1 \sin{\theta_1} \frac{\textrm{d}\theta_1}{\textrm{d}t} \notag
\end{align}
  Here $\frac{\textrm{d}\theta_1}{\textrm{d}t}$ is the angular rate of rotation, earlier denoted as $\omega$.

Next, we take the second derivative:
\begin{align}
    \frac{\textrm{d}^2y_1}{\textrm{d}t^2} =  -R_1 \left[\cos{\theta_1} \left(\frac{\textrm{d}\theta_1}{\textrm{d}t}\right)^2 + \sin{\theta_1}\frac{\textrm{d}^2\theta_1}{\textrm{d}t^2}\right] \notag
\end{align}

Substituting from equation \ref{eq:x1_theta}, and assuming $\frac{\textrm{d}\theta_1}{\textrm{d}t}$ to be constant as before:
\begin{align}
    \ddot y_1 =  -\left( y_1 - o \right) \omega^2 \label{eq:y1}
\end{align}

Equation \ref{eq:y1} also holds for point $P_2$; solving for $o$ we obtain:
\begin{align}
  o &= y_1 + \frac{\ddot y_1}{\omega^2} \notag\\
  o &= y_2 + \frac{\ddot y_2}{\omega^2} \notag
\end{align}

Therefore:
\begin{align}
  y_1 + \frac{\ddot y_1}{\omega^2} &= y_2 + \frac{\ddot y_2}{\omega^2} \notag
\end{align}

Solving for $\omega^2$ we obtain:
\begin{align}
  \omega^2 = -\frac{\ddot y_1 - \ddot y_2}{y_1 - y_2} \notag
\end{align}

Considering that $y = y_1 - y_2$ is the difference between a tracked point and a fixation point, we obtain the same expression as equation \ref{eq:omega}:
\begin{align}
  \omega^2 = -\frac{\ddot y}{y} \notag
\end{align}


\bibliographystyle{IEEEtran}
\bibliography{IEEEabrv,raviv-yepes,invariant}

\end{document}